\documentclass[review]{elsarticle}

\usepackage{graphicx}
\usepackage[tight,normalsize,sf,SF]{subfigure}
\usepackage{booktabs} 
\usepackage{amsmath}
\usepackage{amssymb}
\usepackage{array}
\usepackage{multirow}
\usepackage{color}
\usepackage{ragged2e}
\def\mycolor{\textcolor[rgb]{0, 0, 0}}

\usepackage[pagebackref=false,breaklinks=true,letterpaper=true,colorlinks,bookmarks=false]{hyperref}
\usepackage{enumerate}

\journal{Pattern Recognition}

\begin{document}

\begin{frontmatter}

\title{Hypergraph Convolution and Hypergraph Attention}


\author[label1]{Song Bai\corref{mycorrespondingauthor}}
\ead{songbai.site@gmail.com}
\cortext[mycorrespondingauthor]{Corresponding author}

\author[label1]{Feihu Zhang}
\ead{feihu.zhang@eng.ox.ac.uk}

\author[label1]{Philip H.S. Torr}
\ead{philip.torr@eng.ox.ac.uk}

\address[label1]{Department of Engineering Science, University of Oxford, Oxford, OX1 3PJ, UK}

\begin{abstract}
Recently, graph neural networks have attracted great attention and achieved prominent performance in various research fields. Most of those algorithms have assumed pairwise relationships of objects of interest. However, in many real applications, the relationships between objects are in higher-order, beyond a pairwise formulation. To efficiently learn deep embeddings on the high-order graph-structured data, we introduce two end-to-end trainable operators to the family of graph neural networks, \emph{i.e.}, hypergraph convolution and hypergraph attention. Whilst hypergraph convolution defines the basic formulation of performing convolution on a hypergraph, hypergraph attention further enhances the capacity of representation learning by leveraging an attention module. With the two operators, a graph neural network is readily extended to a more flexible model and applied to diverse applications where non-pairwise relationships are observed. Extensive experimental results with semi-supervised node classification demonstrate the effectiveness of hypergraph convolution and hypergraph attention.
\end{abstract}

\begin{keyword}
Graph Learning, Hypergraph Learning, Graph Neural Networks, Semi-supervised Learning
\end{keyword}

\end{frontmatter}

\section{Introduction}\label{sec:introduction}
In the last decade, Convolution Neural Networks (CNNs)~\cite{krizhevsky2012imagenet} have led to a wide spectrum of breakthrough in various research domains, such as visual recognition~\cite{resnet}, speech recognition~\cite{hinton2012deep}, machine translation~\cite{bahdanau2014neural}~\emph{etc}. Due to its innate nature, CNNs hold an extremely strict assumption, that is, input data shall have a regular and grid-like structure. Such a limitation hinders the promotion and application of CNNs to many tasks where data of irregular structures widely exists.

\begin{figure}[tb]
\centering
\subfigure[]
{
\begin{minipage}[tb]{0.4\textwidth}
\includegraphics[width = 1\textwidth]{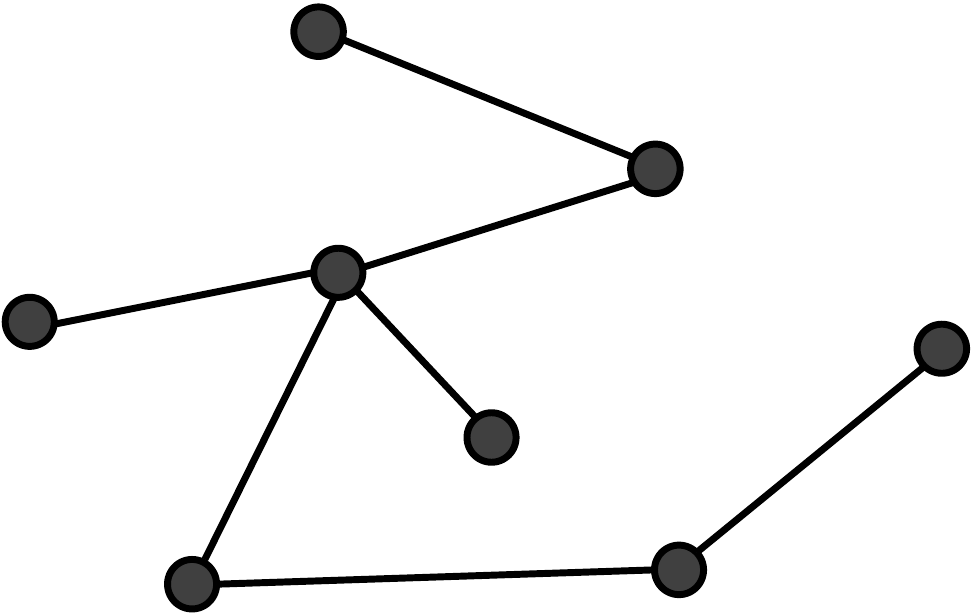}
\end{minipage}
\label{fig:illustration1}
}
\subfigure[]
{
\begin{minipage}[tb]{0.4\textwidth}
\includegraphics[width = 1\textwidth]{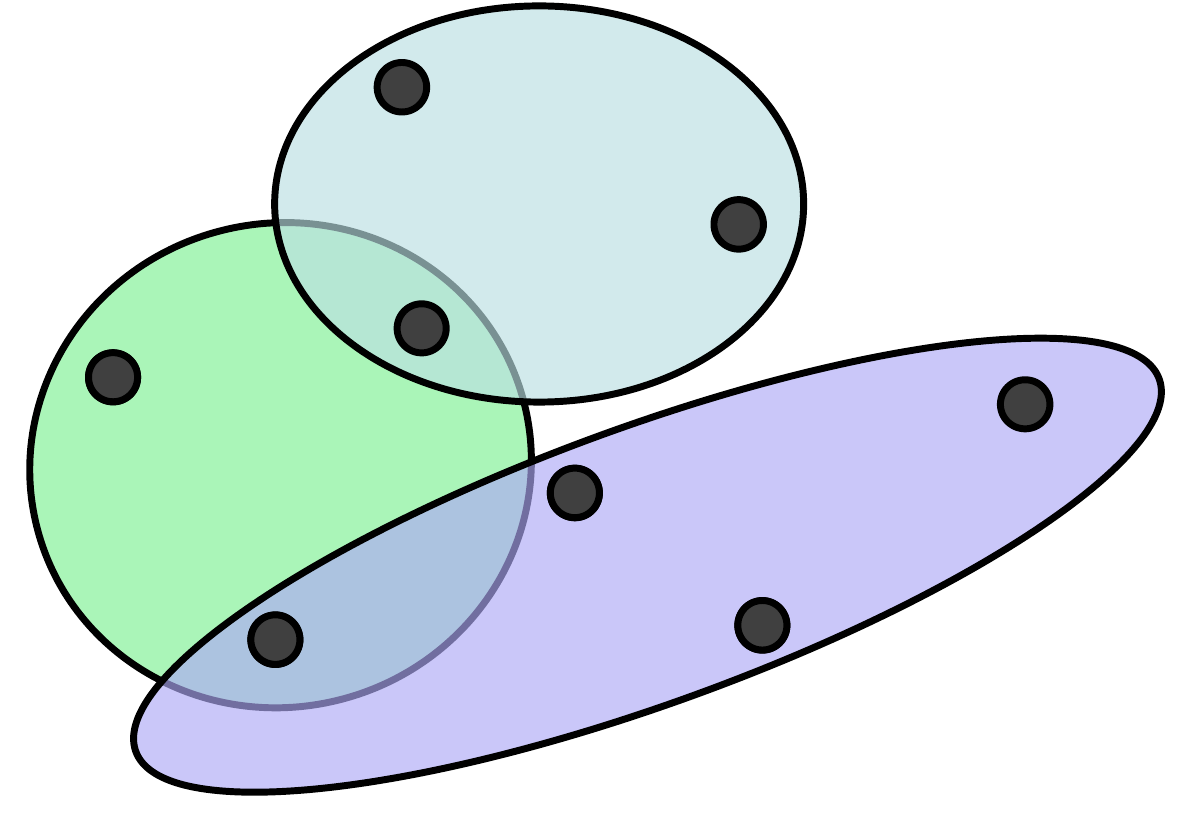}
\end{minipage}
\label{fig:illustration2}
}
\caption{The difference between a simple graph (a) and a hypergraph (b). In a simple graph, each edge, denoted by a line, only connects two vertices. In a hypergraph, each edge, denoted by a colored ellipse, connects more than two vertices.}
\label{fig:illustration}
\end{figure}

To handle the ubiquitous irregular data structures, there is a growing interest in Graph Neural Networks (GNNs)~\cite{scarselli2009graph}, a methodology for learning deep models with graph data. GNNs have a wide application in social science~\cite{hamilton2017inductive}, knowledge graph~\cite{wang2018zero}, recommendation system~\cite{ying2018graph}, geometrical computation~\cite{bronstein2017geometric},~\emph{etc}. And most existing methods assume that the relationships between objects of interest are in pairwise formulations. Specifically in a graph model, it means that each edge only connects two vertices (see Fig.~\ref{fig:illustration1}).

However, in many real applications, the object relationships are much more complex than pairwise. For instance in recommendation systems, an item may be commented by multiple users. By taking the items as vertices and the rating of users as edges of a graph, each edge may connect more than two vertices. In this case, the affinity relations are no longer dyadic (pairwise), but rather triadic, tetradic or of a higher-order. This brings back the concept of hypergraph~\cite{agarwal2006higher,zhang2017re,yu2012adaptive,pedronette2019multimedia}, a special graph model which leverages hyperedges to connect multiple vertices simultaneously (see Fig.~\ref{fig:illustration2}). Unfortunately, most existing variants of graph neural networks~\cite{survey1,survey2} are not applicable to the high-order structure encoded by hyperedges.

Although the machine learning and pattern recognition community has witnessed the prominence of graph neural networks in learning patterns on simple graphs, the investigation of deep learning on hypergraphs is still in a very nascent stage. Considering its importance, we propose hypergraph convolution and hypergraph attention in this work, as two strong supplemental operators to graph neural networks. The advantages and contributions of our work are as follows

\vspace{2ex}\noindent\textbf{1)}~Hypergraph convolution defines a basic convolutional operator in a hypergraph. It enables an efficient information propagation between vertices by fully exploiting the high-order relationship and local clustering structure therein. We mathematically prove that graph convolution is a special case of hypergraph convolution when the non-pairwise relationship degenerates to a pairwise one.

\vspace{2ex}\noindent\textbf{2)}~Apart from hypergraph convolution where the underlying structure used for propagation is pre-defined, hypergraph attention further exerts an attention mechanism to learn a dynamic connection of hyperedges. Then, the information propagation and gathering is done in task-relevant parts of the graph, thereby generating more discriminative node embeddings.

\vspace{2ex}\noindent\textbf{3)}~Both hypergraph convolution and hypergraph attention are end-to-end trainable, and can be inserted into most variants of graph neural networks as long as non-pairwise relationships are observed. Extensive experimental results on benchmark datasets demonstrate the efficacy of the proposed methods for semi-supervised node classification.

The rest of paper is organized as follows. In Sec.~\ref{sec:related}, we briefly review some representative methods in graph learning and graph neural networks. In Sec.~\ref{sec:proposed}, we systematically introduce the motivation, formulation and implementation of the proposed method. Experimental evaluations and comparisons are presented in Sec.~\ref{sec:exp_gnn}, followed by the conclusions and future work given in Sec.~\ref{sec:con}.

\section{Related Work} \label{sec:related}
Graphs are a classic kind of data structure~\cite{yang2012affinity,pedronette2018unsupervised,pedronette2013image}, where its vertices represent objects and its edges linking two adjacent vertices describe the relationship between the corresponding objects.

Graph Neural Network (GNN) is a methodology for learning deep models or embeddings on graph-structured data, which was first proposed by~\cite{scarselli2009graph}. One key aspect in GNN is to define the convolutional operator in the graph domain.~\cite{bruna2013spectral} firstly define convolution in the Fourier domain using the graph Laplacian matrix, and generate non-spatially localized filters with potentially intense computations.~\cite{henaff2015deep} enable the spectral filters spatially localized using a parameterization with smooth coefficients.~\cite{defferrard2016convolutional} focus on the efficiency issue and use a Chebyshev expansion of the graph Laplacian to avoid an explicit use of the graph Fourier basis.~\cite{GCN} further simplify the filtering by only using the first-order neighbors and propose Graph Convolutional Network (GCN), which has demonstrated impressive performance in both efficiency and effectiveness with semi-supervised classification tasks.

Meanwhile, some spatial algorithms directly perform convolution on the graph. For instance,~\cite{duvenaud2015convolutional} learn different parameters for nodes with different degrees, then average the intermediate embeddings over the neighborhood structures.~\cite{niepert2016learning} propose the PATCHY-SAN architecture, which selects a fixed-length sequence of nodes as the receptive field and generate local normalized neighborhood representations for each of the nodes in the sequence.~\cite{atwood2016diffusion} demonstrate that diffusion-based representations can serve as an effective basis for node classification.~\cite{zhuang2018dual} further explore a joint usage of diffusion and adjacency basis in a dual graph convolutional network.~\cite{gilmer2017neural} defines a unified framework via a message passing function, where each vertex sends messages based on its states and updates the states based on the message of its immediate neighbors.~\cite{hamilton2017inductive} propose GraphSAGE, which customizes three aggregating functions,~\emph{i.e.},~element-wise mean, long short-term memory and pooling, to learn embeddings in an inductive setting.

Some other works focus on gate mechanism~\cite{li2015gated}, skip connection~\cite{pham2017column}, jumping connection~\cite{xu2018representation}, attention mechanism~\cite{GAN}, sampling strategy~\cite{chen2018fastgcn,chen2018stochastic}, hierarchical representation~\cite{ying2018hierarchical}, generative models~\cite{you2018graphrnn,bojchevski2018netgan}, adversarial attack~\cite{DBLP:journals/corr/abs-1806-02371},~\emph{etc}. As a thorough review is simply unfeasible due to the space limitation, we refer interested readers to surveys for more representative methods. For example,~\cite{survey1} and~\cite{survey2} present two systematical and comprehensive surveys over a series of variants of graph neural networks.~\cite{bronstein2017geometric} provide a review of geometric deep learning.~\cite{battaglia2018relational} generalize and extend various approaches and show how graph neural networks can support relational reasoning and combinatorial generalization.~\cite{lee2018attention} particularly focus on the attention models for graphs, and introduce three intuitive taxonomies.~\cite{monti2017geometric} propose a unified framework called MoNet, which summarizes Geodesic CNN~\cite{masci2015geodesic}, Anisotropic CNN~\cite{boscaini2016learning}, GCN~\cite{GCN} and Diffusion CNN~\cite{atwood2016diffusion} as its special cases.

As analyzed above, most existing variants of GNN assume pairwise relationships between objects, while our work operates on a high-order hypergraph~\cite{berge1973graphs,li2017inhomogeneous} where the between-object relationships are beyond pairwise. Hypergraph learning methods differ in the structure of the hypergraph,~\emph{e.g.},~clique expansion and star expansion~\cite{zien1999multilevel}, and the definition of hypergraph Laplacians~\cite{bolla1993spectra,rodriguez2003laplacian,zhou2007learning}. Following~\cite{defferrard2016convolutional}, \cite{hypernn} propose a hypergraph neural network using a Chebyshev expansion of the graph Laplacian. By analyzing the incident structure of a hypergraph, our work directly defines two differentiable operators,~\emph{i.e.}, hypergraph convolution and hypergraph attention, which is intuitive and flexible in learning more discriminative deep embeddings.

\section{Proposed Approach} \label{sec:proposed}
In this section, we first give the definition of hypergraph in Sec.~\ref{sec:revisit}, then elaborate the proposed hypergraph convolution and hypergraph attention in Sec.~\ref{sec:hyc} and Sec.~\ref{sec:hya}, respectively. At last, Sec.~\ref{sec:summary} provides a deeper analysis of the properties of our methods.

\subsection{Hypergraph Revisited} \label{sec:revisit}
Most existing works~\cite{GCN,GAN} operate on a simple graph $\mathcal{G}=(V,E)$, where $V=\{v_1,v_2,...,v_N\}$ denotes the vertex set and $E\subseteq V\times V$ denotes the edge set. A graph adjacency matrix $\mathbf{A}\in\mathbb{R}^{N\times N}$ is used to reflect the pairwise relationship between every two vertices. The underlying assumption of such a simple graph is that each edge only links two vertices. However, as analyzed above, the relationships between objects are more complex than pairwise in many real applications. 

To describe such a complex relationship, a useful graph model is hypergraph, where a hyperedge can connect more than two vertices. Let $\mathcal{G}=(V,E)$ be a hypergraph with $N$ vertices and $M$ hyperedges. Each hyperedge $\epsilon\in E$ is assigned a positive weight $W_{\epsilon\epsilon}$, with all the weights stored in a diagonal matrix $\mathbf{W}\in\mathbb{R}^{M\times M}$. Apart from a simple graph where an adjacency matrix is defined, the hypergraph $\mathcal{G}$ can be represented by an incidence matrix $\mathbf{H}\in\mathbb{R}^{N\times M}$ in general. When the hyperedge $\epsilon\in E$ is incident with a vertex $v_i\in V$, in other words, $v_i$ is connected by $\epsilon$, $H_{i\epsilon}=1$, otherwise $0$. Then, the vertex degree is defined as
\begin{equation}
D_{ii}=\sum_{\epsilon=1}^MW_{\epsilon\epsilon}H_{i\epsilon}
\end{equation}
and the hyperedge degree is defined as
\begin{equation}
B_{\epsilon\epsilon}=\sum_{i=1}^NH_{i\epsilon}.
\end{equation}
Note that $\mathbf{D}\in\mathbb{R}^{N\times N}$ and $\mathbf{B}\in\mathbb{R}^{M\times M}$ are both diagonal matrices.

In the following, we define the operator of convolution on the hypergraph $\mathcal{G}$.

\subsection{Hypergraph Convolution} \label{sec:hyc}
The primary obstacle to defining a convolution operator in a hypergraph is to measure the transition probability between two vertices, with which the embeddings (or features) of each vertex can be propagated in a graph neural network. To achieve this, we hold two assumptions: 1) more propagations should be done between those vertices connected by a common hyperedge, and 2) the hyperedges with larger weights deserve more confidence in such a propagation. Then, one step of hypergraph convolution is defined as
\begin{equation} \label{eq:hc1}
x_i^{(l+1)}=\sigma\left(\sum_{j=1}^N\sum_{\epsilon=1}^MH_{i\epsilon}H_{j\epsilon}W_{\epsilon\epsilon}x_j^{(l)}\mathbf{P}\right),
\end{equation}
where $x_i^{(l)}$ is the embedding of the $i$-th vertex in the $(l)$-th layer.  $\sigma(\cdot)$ is a non-linear activation function like LeakyReLU~\cite{leaky} and eLU~\cite{elu}.~$\mathbf{P}\in\mathbb{R}^{F^{(l)}\times F^{(l+1)}}$ is the weight matrix between the $(l)$-th and $(l+1)$-th layer. Eq.~\eqref{eq:hc1} can be written in a matrix form as 
\begin{equation} \label{eq:hc2}
\mathbf{X}^{(l+1)}=\sigma(\mathbf{HW}\mathbf{H}^\mathrm{T}\mathbf{X}^{(l)}\mathbf{P}),
\end{equation}
where $\mathbf{X}^{(l)}\in\mathbb{R}^{N\times F^{(l)}}$ and $\mathbf{X}^{(l+1)}\in\mathbb{R}^{N\times F^{(l+1)}}$ are the input of the $(l)$-th and $(l+1)$-th layer, respectively.

However, $\mathbf{HW}\mathbf{H}^\mathrm{T}$ does not hold a constrained spectral radius, which means that the scale of $\mathbf{X}^{(l)}$ will be possibly changed. In optimizing a neural network, stacking multiple hypergraph convolutional layers like Eq.~\eqref{eq:hc2} can then lead to numerical instabilities and increase the risk of exploding/vanishing gradients. Therefore, a proper normalization is necessary. Thus, we impose a symmetric normalization and arrive at our final formulation
\begin{equation} \label{eq:hc3}
\mathbf{X}^{(l+1)}=\sigma(\mathbf{D}^{-1/2}\mathbf{HW}\mathbf{B}^{-1}\mathbf{H}^\mathrm{T}\mathbf{D}^{-1/2}\mathbf{X}^{(l)}\mathbf{P}).
\end{equation}
Here, we recall that $\mathbf{D}$ and $\mathbf{B}$ are the degree matrices of the vertex and hyperedge in a hypergraph, respectively.
It is easy to prove that the maximum eigenvalue of  $\mathbf{D}^{-1/2}\mathbf{HW}\mathbf{B}^{-1}\mathbf{H}^\mathrm{T}\mathbf{D}^{-1/2}$ is no larger than $1$, which stems from a fact~\cite{agarwal2006higher} that $\textbf{I}-\mathbf{D}^{-1/2}\mathbf{HW}\mathbf{B}^{-1}\mathbf{H}^\mathrm{T}\mathbf{D}^{-1/2}$ is a positive semi-definite matrix. $\mathbf{I}$ is an identity matrix of an appropriate size.

Alternatively, a row-normalization is also viable as
\begin{equation} \label{eq:hc4}
\mathbf{X}^{(l+1)}=\sigma(\mathbf{D}^{-1}\mathbf{HW}\mathbf{B}^{-1}\mathbf{H}^\mathrm{T}\mathbf{X}^{(l)}\mathbf{P}),
\end{equation}
which enjoys similar mathematical properties as Eq.~\eqref{eq:hc3}, except that the propagation is directional and asymmetric in this case.

As $\mathbf{X}^{(l+1)}$ is differentiable with respect to $\mathbf{X}^{(l)}$ and $\mathbf{P}$, we can use hypergraph convolution in model training and optimize it via gradient descent.

\subsection{Hypergraph Attention} \label{sec:hya}
\mycolor{Hypergraph convolution has a sort of innate attentional mechanism~\cite{lee2018attention,GAN}. As we can find from Eq.~\eqref{eq:hc3} and Eq.~\eqref{eq:hc4}, the transition probability between vertices is non-binary, which means that for a given vertex, the afferent and efferent information flow is explicitly assigned a diverse magnitude of importance. However, such an attentional mechanism is not learnable and trainable after the graph structure (the incidence matrix $\mathbf{H}$) is given. The goal of hypergraph attention is to learn a dynamic incidence matrix, thereby a dynamic transition matrix that can better reveal the intrinsic relationship between vertices.}

One natural solution is to exert an attention learning module on $\mathbf{H}$. In this circumstance, instead of treating each vertex as being connected by a certain hyperedge or not, the attention module presents a probabilistic model, which assigns non-binary and real values to measure the degree of connectivity. We expect that the probabilistic model can learn more category-discriminative embeddings and the relationship between vertices can be more accurately described.

Nevertheless, hypergraph attention is only feasible when the vertex set and the hyperedge set are from (or can be projected to) the same homogeneous domain, since only in this case, the similarities between vertices and hyperedges are directly comparable. In practice, it depends on how the hypergraph $\mathcal{G}$ is constructed. For example,~\cite{huang2010image} apply hypergraph learning to image retrieval where each vertex collects its k-nearest neighbors to form a hyperedge, as also the way of constructing hypergraphs in our experiments. When the vertex set and the edge set are comparable, we define the procedure of hypergraph attention inspired by~\cite{GAN}. For a given vertex $x_i$ and its associated hyperedge $x_j$, the attentional score is 
\begin{equation} \label{eq:ha1}
H_{ij}=\frac{\exp\left(\sigma(\text{sim}(x_i\mathbf{P}, x_j\mathbf{P}))\right)}{\sum_{k\in\mathcal{N}_i}\exp\left(\sigma(\text{sim}(x_i\mathbf{P}, x_k\mathbf{P}))\right)},
\end{equation}
where $\sigma(\cdot)$ is a non-linear activation function. $\mathcal{N}_i$ is the neighborhood set of $x_i$, which can be pre-accessed on some benchmarks, such as the Cora and Citeseer datasets~\cite{cora}. sim$(\cdot)$ is a similarity function that computes the pairwise similarity between two vertices, defined as
\begin{equation}
\text{sim}(x_i, x_j)=\textbf{a}^\mathrm{T}[x_i\|x_j]. 
\end{equation}
Here $[,\|,]$ denotes concatenation and $\textbf{a}$ is a weight vector used to output a scalar similarity value.

With the incidence matrix $\mathbf{H}$ enriched by an attention module, one can also follow Eq.~\eqref{eq:hc3} and Eq.~\eqref{eq:hc4} to learn the intermediate embedding of vertices layer-by-layer. Note that hypergraph attention also propagates gradients to $\mathbf{H}$ in addition to $\mathbf{X}^{(l)}$ and $\mathbf{P}$. 
 
In some applications, the vertex set and the hyperedge set are from two heterogeneous domains. For instance,~\cite{zhou2007learning} assume that attributes are hyperedges to connect objects like newspaper or text. Then, it is problematic to directly learn an attention module over the incidence matrix $\mathbf{H}$. We leave this issue for future work.

\subsection{Summary and Remarks} \label{sec:summary}
\begin{figure}[tb]
\centering
\includegraphics[width=0.9\linewidth]{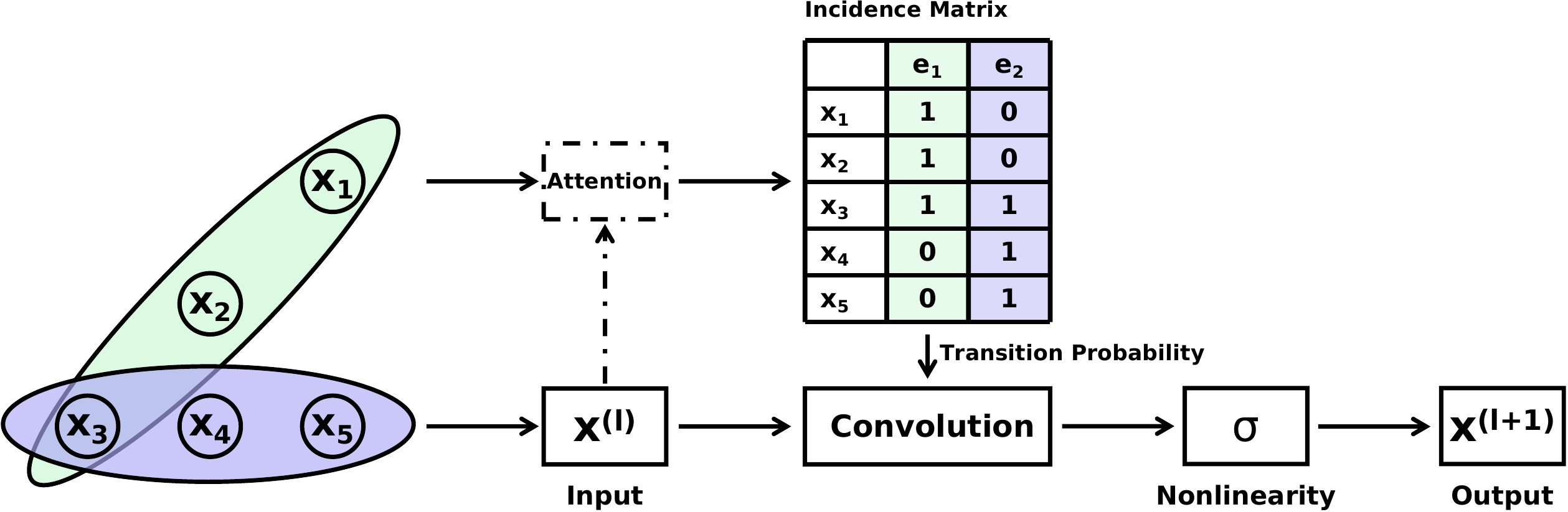}
\caption{Schematic illustration of hypergraph convolution with $5$ vertices and $2$ hyperedges. With an optional attention mechanism, hypergraph convolution upgrades to hypergraph attention.}
\label{fig:pipeline}
\end{figure}

The pipeline of the proposed hypergraph convolution and hypergraph attention is illustrated in Fig.~\ref{fig:pipeline}. Both two operators can be inserted into most variants of graph neural networks when non-pairwise relationships are observed, and used for model training. 

As the only difference between hypergraph convolution and hypergraph attention is an optional attention module on the incidence matrix $\mathbf{H}$, below we take hypergraph convolution as a representative to further analyze the properties of our methods. Note that the analyses also hold for hypergraph attention.

\vspace{2ex}\noindent\textbf{Relationship with Graph Convolution.}~We prove that graph convolution~\cite{GCN} is a special case of hypergraph convolution mathematically.

Let $\mathbf{A}\in\mathbb{R}^{N\times N}$ be the adjacency matrix used in graph convolution. When each edge only links two vertices in a hypergraph, the vertex degree matrix $\mathbf{B}=2\mathbf{I}$. Assuming equal weights for all the hyperedges (\emph{i.e.}, $\mathbf{W}=\mathbf{I}$), we have an interesting observation of hypergraph convolution. Based on Eq.~\eqref{eq:hc3}, the definition of hypergraph convolution then becomes
\begin{equation} \label{eq:summmary1}
\begin{split}
\mathbf{X}^{(l+1)}=&\sigma(\frac12\mathbf{D}^{-1/2}\mathbf{H}\mathbf{H}^\mathrm{T}\mathbf{D}^{-1/2}\mathbf{X}^{(l)}\mathbf{P}), \\
=&\sigma\left(\frac12\mathbf{D}^{-1/2}(\mathbf{A}+\mathbf{D})\mathbf{D}^{-1/2}\mathbf{X}^{(l)}\mathbf{P}\right)\\
=&\sigma\left(\frac12(\mathbf{I}+\mathbf{D}^{-1/2}\mathbf{A}\mathbf{D}^{-1/2})\mathbf{X}^{(l)}\mathbf{P}\right) \\
=&\sigma(\hat{\mathbf{A}}\mathbf{X}^{(l)}\mathbf{P}),
\end{split}
\end{equation}
where $\hat{\mathbf{A}}=1/2\widetilde{\mathbf{A}}$ and $\widetilde{\mathbf{A}}=\mathbf{I}+\mathbf{D}^{-1/2}\mathbf{A}\mathbf{D}^{-1/2}$. As we can see, Eq.~\eqref{eq:summmary1} is exactly equivalent to the definition of graph convolution (see Equation 9 in~\cite{GCN}). Note that $\widetilde{\mathbf{A}}$ has eigenvalues in the range $[0,2]$. To avoid scale changes,~\cite{GCN} have suggested a re-normalization trick, that is

\begin{equation} \label{eq:xx}
\hat{\mathbf{A}}=\mathbf{\widetilde{D}}^{-1/2}\widetilde{\mathbf{A}}\mathbf{\widetilde{D}}^{-1/2}.
\end{equation}
In the specific case of hypergraph convolution, we are using a simplified solution, that is dividing $\widetilde{\mathbf{A}}$ by $2$.

With GCN as a bridge and springboard to the family of graph neural networks, it then becomes feasible to build connections with other frameworks,~\emph{e.g.}, MoNet \cite{monti2017geometric}, and develop the higher-order counterparts of those variants to deal with non-pairwise relationships.

\vspace{2ex}\noindent\textbf{Implementation in Practice and \mycolor{Complexity Analysis.}}~The implementation of hypergraph convolution appears sophisticated as $6$ matrices are multiplied for symmetric convolution (see Eq.~\eqref{eq:hc3}) and $5$ matrices are multiplied for asymmetric convolution (see Eq.~\eqref{eq:hc4}). However, it should be mentioned that $\textbf{D}$, $\textbf{W}$ and $\textbf{B}$ are all diagonal matrices, which makes it possible to efficiently implement it in commonly used deep learning platforms.

For asymmetric convolution, we have from Eq.~\eqref{eq:hc4} that
\begin{equation} \label{eq:remark2}
\mathbf{D}^{-1}\mathbf{HW}\mathbf{B}^{-1}\mathbf{H}^\mathrm{T}=(\mathbf{D}^{-1}\mathbf{H}\textbf{W})(\mathbf{H}\mathbf{B}^{-1})^\mathrm{T},
\end{equation}
where $\mathbf{D}^{-1}\mathbf{HW}$ performs $L_1$ normalization of $\mathbf{HW}$ over rows and $\mathbf{H}\mathbf{B}^{-1}$ performs $L_1$ normalization of $\mathbf{H}$ over columns. In space-saving applications where matrix-form variables are allowed, normalization can be simply done using standard built-in functions in public neural network packages.

In case of space-consuming applications, one can readily implement a sparse version of hypergraph convolution as well. Since $\mathbf{H}$ is usually a sparse matrix, Eq.~\eqref{eq:remark2} does not necessarily decrease the sparsity too much. Hence, we can conduct normalization only on non-zero indices of $\mathbf{H}$, resulting in a sparse transition matrix. 

\mycolor{Mathematically, if we implement Eq.~\eqref{eq:remark2} directly via matrix multiplications, the time complexity would be upper bounded by O($2N^2M+2M^2N$). In comparison, if implemented via normalization, the time complexity is significantly decreased. In more detail, $\mathbf{D}^{-1}\mathbf{HW}$ requires $2NM$ operations, $\mathbf{H}\mathbf{B}^{-1}$ requires $NM$ operations, and the matrix multiplication between them requires $N^2M$ operations. 
As a result, the total operations required are bounded by O($3NM+N^2M$). It can be envisioned that the speed can be even faster if the matrix sparsity is considered. Because of the well-optimized normalization functions in commonly used computing platforms (\emph{e.g.,}~\texttt{torch.nn.functional.normalize} in Pytorch or \texttt{sklearn.preprocessing.normalize} in scikit-learn), it only takes $1.8$ms to fulfill one forward pass on the Cora dataset using a server with an Intel(R) Core(TM) i7-5960X CPU (3.00GHz) and an NVIDIA GeForce RTX 2080 Ti GPU.}


Symmetric hypergraph convolution defined in Eq.~\eqref{eq:hc3} can be implemented similarly, with a minor difference in normalization using the vertex degree matrix $\mathbf{D}$.

\vspace{2ex}\noindent\textbf{Skip Connection.}~Hypergraph convolution can be integrated with skip connection~\cite{resnet} as
\begin{equation} \label{eq:xx}
\mathbf{X}^{(l+1)}_k=\text{HConv}\left(\mathbf{X}^{(l)},\mathbf{H}_k,\mathbf{P}_k\right)+\mathbf{X}^{(l)},
\end{equation}
where HConv$(\cdot)$ represents the hypergraph convolution operator defined in Eq.~\eqref{eq:hc3} (or Eq.~\eqref{eq:hc4}). Some similar structures (\emph{e.g.}, highway structure~\cite{srivastava2015highway} adopted in Highway-GCN~\cite{rahimi2018semi}) can be also applied. 

It has been demonstrated~\cite{GCN} that deep graph models cannot improve the performance even with skip connections since the receptive field grows exponentially with respect to the model depth. In the experiments, we will verify the compatibility of the proposed operators with skip connections in model training.

\vspace{2ex}\noindent\textbf{Multi-head.}~To stabilize the learning process and improve the representative power of networks, multi-head (\emph{a.k.a.}~multi-branch) architecture is suggested in relevant works,~\emph{e.g.},~\cite{resnext,vaswani2017attention,GAN}. hypergraph convolution can be also extended in that way, as
\begin{equation} \label{eq:xx}
\begin{split}
\mathbf{X}^{(l+1)}_k&=\text{HConv}\left(\mathbf{X}^{(l)},\mathbf{H}_k,\mathbf{P}_k\right), \\
\mathbf{X}^{(l+1)}&=\text{Aggregate}\left(\mathbf{X}^{(l+1)}_k\right)_{k=1}^K,
\end{split}
\end{equation}
where Aggregate$(\cdot)$ is a certain aggregation like concatenation or average pooling. $\mathbf{H}_k$ and $\mathbf{P}_k$ are the incidence matrix and weight matrix corresponding to the $k$-th head, respectively. Note that only in hypergraph attention, $\mathbf{H}_k$ is different over different heads.

\section{Experiments} \label{sec:exp_gnn}
In this section, we evaluate the proposed hypergraph convolution and hypergraph attention in the task of semi-supervised node classification.

Following~\cite{GCN,GAN}, we first employ three citation network datasets, including the Cora, Citeseer and Pubmed datasets~\cite{cora}, to make a fair comparison with previous methods.
\begin{itemize}
	\item The Cora dataset contains $2,708$ scientific publications divided into $7$ categories. There are $5,429$ edges in total, with each edge being a citation link from one article to another. Each publication is described by a binary bag-of-word representation, where $0$ (or $1$) indicates the absence (or presence) of the corresponding word from the dictionary. The dictionary consists of $1,433$ unique words.
	\item Like the Cora dataset, the Citeseer dataset contains $3,327$ scientific publications, divided into $6$ categories and linked by $4,732$ edges. Each publication is described by a binary bag-of-word representation of $3,703$ dimensions.
	\item The Pubmed dataset is comprised of $19,717$ scientific publications divided into $3$ classes. The citation network has $44,338$ links. Each publication is described by a vectorial representation using Term Frequency-Inverse Document Frequency (TF-IDF), drawn from a dictionary with $500$ terms.
\end{itemize}

Then as per~\cite{zhou2007learning}, a modified version of the 20-newsgroup dataset\footnote{\url{https://cs.nyu.edu/~roweis/data.html}} with binary occurrence values for 100 words is used for text categorization. It consists of $16,242$ articles divided into $4$ groups, with the sizes being $4605$, $3519$, $2657$ and $5461$ respectively. Each word naturally connects multiple postings, which makes this dataset more suitable for the use of a hypergraph. In the meantime, it means the graph used is generated using the same resources (words) as the node features. Table~\ref{Table:datasets} presents an overview of the dataset statistics.

As for the training-testing data split, we adopt the setting used in~\cite{yang2016revisiting}. In each dataset, $20$ nodes per category are used for model training. Another $500$ nodes are used for validation purposes and $1000$ nodes are used for performance evaluation.

\begin{table}[tb]
\small
\centering
\begin{tabular}{|l|*{4}{p{1.28cm}<{\centering}}|}
\hline
Dataset & \#Nodes  & \#Edges & \#Features & \#Classes  \\
\hline
\hline
Cora & 2708 & 5429 & 1433 &  7 \\
Citeseer & 3327  & 4732 & 3703 & 6 \\
Pubmed & 19717 & 44338 & 500 &  3 \\
20-newsgroup & 16242 & - & 100 & 4 \\
\hline
\end{tabular}
\caption{Overview of data statistics.}
\label{Table:datasets}
\end{table}

\subsection{Experiments on Citation Networks} \label{sec:exp_citation}

\subsubsection{Hypergraph Construction}

Most existing methods interpret the citation network as the adjacency matrix of a simple graph by a certain kind of normalization,~\emph{e.g.},~\cite{GCN}. By doing so, these methods working on a simple graph aim to define the scheme of message passing if one article is cited by another article.

In this work, we construct a higher-order graph to enable hypergraph convolution and hypergraph attention. The whole procedure is divided into three steps: 1) all the articles constitute the vertex set of the hypergraph; 2) each article is taken as a centroid and forms a hyperedge to connect those articles which have citation links to it (either citing it or being cited); 3) the hyperedges are equally weighted for simplicity, but one can set non-equal weights to encode a prior knowledge if existing in other applications. Compared with existing methods, our work focuses on the message passing if two articles are both cited by a third article.

\subsubsection{Implementation Details} \label{sec:exp_setup}

We implement the proposed hypergraph convolution and hypergraph attention using Pytorch\footnote{Included in the PyTorch Geometric Library:~\url{https://github.com/rusty1s/pytorch_geometric}}. As for the parameter setting and network structure, we closely follow~\cite{GAN} without a carefully parameter tuning and model design. 

In more detail, a two-layer graph model is constructed. The first layer consists of $8$ branches of the same topology, and each branch generates an $8$-dimensional hidden representation. The second layer, used for classification, is a single-branch topology and generates $C$-dimensional feature ($C$ is the number of classes). Each layer is followed by a nonlinearity activation and here we use Exponential Linear Unit (ELU)~\cite{elu}. $L_2$ regularization is applied to the parameters of network with $\lambda=0.0003$ on the Cora and Citeseer datasets and $\lambda=0.001$ on the Pubmed dataset, respectively.

Specifically in hypergraph attention, dropout with a rate of $0.6$ is applied to both inputs of each layer and the attention transition matrix. As for the computation of the attention incidence matrix $\mathbf{H}$ in Eq.~\eqref{eq:ha1}, we employ a linear transform as the similarity function sim$(\cdot)$, followed by LeakyReLU nonlinearity~\cite{leaky} with the negative input slope set to $0.2$. On the Pubmed dataset, we do not use $8$ output attention heads for classification to ensure the consistency of network structures.

We train the model by minimizing the cross-entropy loss on the training nodes using the Adam~\cite{adam} optimizer with a learning rate of $0.005$ on the Cora and Citeseer datasets and $0.01$ on the Pubmed dataset, respectively. An early stop strategy is adopted on the validation loss with a patience of $100$ epochs. For all the experiments, we report the mean classification accuracy and standard deviation of $100$ trials on the testing dataset. 


\subsubsection{Analysis} \label{sec:analysis}
We first analyze the properties of hypergraph convolution and hypergraph attention with a series of ablation studies. The comparison is primarily done with Graph Convolution Network (GCN)~\cite{GCN} and Graph Attention Network (GAT)~\cite{GAN}, which are two latest representatives of graph neural networks that have close relationships with our methods.

For a fair comparison, we reproduce the performance of GCN and GAT with \emph{exactly} the same experimental setting aforementioned. Thus, we denote them by GCN* and GAT* in the following. Moreover, we employ the same normalization strategy as GCN,~\emph{i.e.},~symmetric normalization in Eq.~\eqref{eq:hc3} for hypergraph convolution, and the same strategy as GAT,~\emph{i.e.},~asymmetric normalization in Eq.~\eqref{eq:hc4} for hypergraph attention. They are denoted by Hyper-Conv. and Hyper-Atten. for short, respectively. 

We modify the model of GAT to implement GCN by removing the attention module and directly feeding the graph adjacency matrix with the normalization trick proposed in GCN. Two noteworthy comments are made here. First, although the architecture of GCN* differs from the original one, the principle of performing graph convolution is the same. Second, directly feeding the graph adjacency matrix is not equivalent to the constant attention described in GAT as normalization is used in our case. In GAT, the constant attention weight is set to $1$ without normalization.

\vspace{2ex}\noindent\textbf{Comparisons with Baselines.}~The comparison with baseline methods is given in Table~\ref{Table:baseline}. 
\begin{table}[tb]
\small
\centering
\begin{tabular}{|l|cc|}
\hline
Method & Cora dataset & Citeseer dataset \\
\hline
\hline
GCN* & 81.80 & 70.29  \\
Hyper-Conv. (\textbf{ours}) & 82.19 & \textbf{70.35} \\
GCN*+Hyper-Conv. (\textbf{ours}) & \textbf{82.63} & 70.00 \\
\hline
\hline
GAT* & 82.43 & 70.02 \\
Hyper-Atten. (\textbf{ours})  & 82.61 & \textbf{70.88}  \\
GAT*+Hyper-Atten. (\textbf{ours})  & \textbf{82.74} & 70.12  \\
\hline
\end{tabular}
\caption{The comparison with baseline methods in terms of classification accuracy (\%). ``Hyper-Conv." denotes hypergraph convolution and ``Hyper-Atten." denotes hypergraph attention.}
\label{Table:baseline}
\end{table}

We first observe that hypergraph convolution and hypergraph attention, as \textbf{non-pairwise models}, consistently outperform its corresponding \textbf{pairwise models},~\emph{i.e.},~graph convolution network (GCN*) and graph attention network (GAT*). For example on the Cora dataset, GCN* achieves an accuracy of $81.80$ and hypergraph convolution reports an accuracy of $82.19$. We conduct onesided two-sample t-test and obtain p-value equal to $0.016$, indicating the improvement is statistically significant at the $5\%$ significance level. On the Citeseer dataset, hypergraph attention achieves a classification accuracy of $70.88$, an improvement of $0.86$ over GAT*. This demonstrates the benefit of considering higher-order models in graph neural networks to exploit non-pairwise relationships and local clustering structure parameterized by hyperedges.

Compared with hypergraph convolution, hypergraph attention adopts a data-driven learning module to dynamically estimate the strength of each link associated with vertices and hyperedges. Thus, the attention mechanism helps hypergraph convolution embed the non-pairwise relationships between objects more accurately. As presented in Table~\ref{Table:baseline}, the performance improvements brought by hypergraph attention are $0.42$ and $0.53$ over hypergraph convolution on the Cora and Citeseer datasets, respectively.

Although non-pairwise models proposed in this work have achieved improvements over pairwise models, one cannot hastily deduce that non-pairwise models are more capable in learning robust deep embeddings under all circumstances. A rational claim is that they are suitable for different applications as real data may convey different structures. Some graph-structured data can be only modeled in a simple graph, some can be only modeled in a higher-order graph and others are suitable for both. Nevertheless, as analyzed in Sec.~\ref{sec:summary}, our method presents a more flexible operator in graph neural networks, where graph convolution and graph attention are special cases of non-pairwise models with guaranteed mathematical properties and performance.

One may be also interested in another question,~\emph{i.e.},~does it bring performance improvements if using hypergraph convolution (or attention) in conjunction with graph convolution (or attention)? We further investigate this by averaging the transition probability learned by non-pairwise models and pairwise models with equal weights, and report the results in Table~\ref{Table:baseline}. As it shows, a positive synergy is only observed on the Cora dataset, where the best results of convolution operator and attention operator are improved to $82.63$ and $82.74$, respectively. By contrast, our methods encounter a slight performance decrease on the Citeseer dataset. From another perspective, it also supports our above claim that different data may fit different structural graph models.

\vspace{2ex}\noindent\textbf{Analysis of Skip Connection.}~We study the influence of skip connection~\cite{resnet} by adding an identity mapping in the first hypergraph convolution layer. We report in Table~\ref{Table:skip} two settings of the weight decay,~\emph{i.e.},~$\lambda$=3e-4 (default setting) and $\lambda$=1e-3.
\begin{table}[tb]
\small 
\centering
\begin{tabular}{|l|cc|cc|}
\hline
\multirow{2}{*}{Method} & \multicolumn{2}{c|}{$\lambda$=3e-4} & \multicolumn{2}{c|}{$\lambda$=1e-3} \\
\cline{2-3} \cline{4-5}
& Cora  & Citeseer & Cora  & Citeseer  \\
\hline
\hline
GCN*           & 79.96 & 69.24 & 80.52 & 70.15  \\
Hyper-Conv. (\textbf{ours})   & 82.22 & 69.46 & 82.66 & 70.83  \\
GAT*           & 80.84 & 68.96 & 81.33 & 69.69  \\
Hyper-Atten. (\textbf{ours})  & \textbf{81.85} & \textbf{70.37} & \textbf{82.34} & \textbf{71.19}  \\
\hline
\end{tabular}
\caption{The compatibility of skip connection in terms of accuracy (\%).}
\label{Table:skip}
\end{table}

As it shows, both GCN* and GAT* report lower recognition rates when integrated with skip connection compared with those reported in Table~\ref{Table:baseline}. In comparison, the proposed non-pairwise models, especially hypergraph convolution, seem to benefit from skip connection. For instance, the best-performing trial of hypergraph convolution yields $82.66$ on the Cora dataset, better than $82.19$ achieved without skip connection, and yields $70.83$ on the Citeseer dataset, better than $70.35$ achieved without skip connection.

Such experimental results encourage us to further train a much deeper model implemented with hypergraph convolution or hypergraph attention (say up to $10$ layers). However, we also witness a performance deterioration either with or without skip connection. This reveals that a better training paradigm and architecture are still urgently required for graph neural networks.

\vspace{2ex}\noindent\textbf{Analysis of Hidden Representation.}~Table~\ref{Table:hidden} presents the performance comparison between GCN* and hypergraph convolution with an increasing length of the hidden representation. The number of heads is set to $1$.
\begin{table}[tb]
\small
\centering

\begin{tabular}{|l|cccccc|}
\hline
\multirow{2}{*}{Method} & \multicolumn{6}{c|}{The length of hidden representation}\\
\cline{2-7}
           & 2 & 4 & 8 & 16 & 24 & 36 \\
\hline
\hline
GCN*          & 65.9 & 79.6 & \textbf{82.0} & 81.9 & 82.0 & 81.9 \\
Hyper-Conv.  & \textbf{69.7} & \textbf{80.4} & \textbf{82.0} & \textbf{82.1} & \textbf{82.1} & \textbf{82.1} \\
\hline
\end{tabular}
\caption{The influence of the length of hidden representation on Cora.}
\label{Table:hidden}
\end{table}

It is easy to find that the performance keeps increasing with an increase of the length of the hidden representation, then peaks when the length is $16$. Moreover, hypergraph convolution consistently beats GCN* with a variety of feature dimensions. As the only difference between GCN* and hypergraph convolution is the used graph structure, the performance gain purely comes from a more robust way of establishing the relationships between objects. It firmly demonstrates the ability of our methods in graph knowledge embedding.

\begin{table*}[tb]
\small
\centering
\begin{tabular}{|l|ccc|}
\hline
Method & Cora & Citeseer & Pubmed  \\
\hline
\hline
Multilayer Perceptron & 55.1 & 46.5 & 71.4 \\
Manifold Regularization~\cite{belkin2006manifold} & 59.5 & 60.1 & 70.7 \\
Semi-supervised Embedding~\cite{weston2012deep} & 59.0 & 59.6 & 71.7 \\
Label Propagation~\cite{LP} & 68.0  & 45.3 &  63.0 \\
DeepWalk~\cite{perozzi2014deepwalk} & 67.2 & 43.2 & 65.3 \\
Iterative Classification Algorithm~\cite{lu2003link} & 75.1 & 69.1 & 73.9 \\
Planetoid~\cite{yang2016revisiting} & 75.7 & 64.7 & 77.2 \\
Chebyshev~\cite{defferrard2016convolutional} & 81.2 & 69.8 & 74.4 \\
Graph Convolutional Network~\cite{GCN} & 81.5 & 70.3 & \textbf{\color{red}{79.0}} \\
Feng~\emph{et al.}~\cite{hypernn} & 81.6 & - & - \\
MoNet~\cite{monti2017geometric} & 81.7 & - & \textbf{\color{blue}{78.8}} \\
Variance Reduction~\cite{chen2018stochastic} & 82.0 & 70.9 & \textbf{\color{red}{79.0}} \\
Graph Attention Network~\cite{GAN} & \textbf{\color{red}{83.0}} & \textbf{\color{red}{72.5}} & \textbf{\color{red}{79.0}} \\
\textbf{Ours}   & \textbf{\color{blue}{82.7$\pm0.3$}} & \textbf{\color{blue}{71.2$\pm0.4$}}  & 78.4$\pm$0.3 \\
\hline
\end{tabular}
\caption{Comparison with the state-of-the-art methods in terms of classification accuracy (\%). The best and second best results are marked in red and blue, respectively.}
\label{Table:sota}
\end{table*}

\subsubsection{Comparison with State-of-the-art}
We compare our method with the state-of-the-art algorithms, which have followed the experimental setting in~\cite{yang2016revisiting} and reported classification accuracies on the Cora, Citeseer and Pubmed datasets. Note that the results are directly quoted from the original papers, instead of being re-implemented in this work. Besides GCN and GAT, the selected algorithms also include Manifold Regularization~\cite{belkin2006manifold} , Semi-supervised Embedding~\cite{weston2012deep}, Label Propagation~\cite{LP}, DeepWalk~\cite{perozzi2014deepwalk}, Iterative Classification Algorithm~\cite{lu2003link}, Planetoid~\cite{yang2016revisiting}, Chebyshev~\cite{defferrard2016convolutional}, MoNet~\cite{monti2017geometric}, and Variance Reduction~\cite{chen2018stochastic}.

As presented in Table~\ref{Table:sota}, our method achieves the second best performance on the Cora and Citeseer datasets, which is slightly inferior to GAT~\cite{GAN} by $0.3$ and $1.3$, respectively. The performance gap is attributed to multiple factors, such as the difference in deep learning platforms and better parameter tuning. As shown in Sec.~\ref{sec:analysis}, thorough experimental comparisons under the same setting have demonstrated the benefit of learning deep embeddings using the proposed non-pairwise models. Nevertheless, we emphasize again that pairwise and non-pairwise models have different application scenarios, and existing variants of graph neural networks can be easily extended to their non-pairwise counterparts with the proposed two operators.
 
On the Pubmed dataset, hypergraph attention reports a classification accuracy of $78.4$, better than $78.1$ achieved by GAT*. As described in Sec.~\ref{sec:exp_setup}, the original implementation of GAT adopts $8$ output attention heads while only $1$ is used in GAT* to ensure the consistency of the model architecture. Even though, hypergraph attention also achieves a comparable performance with the state-of-the-art methods.

\subsection{Experiments on Text Categorization}
In the experiments presented in Sec.~\ref{sec:exp_citation}, we assume that \mycolor{a pre-defined graph structure, independent from the input features, is given. For example, the citation links presented in the citation network are irrelevant to the feature embeddings of articles.}
Here, we show the scenario where graph structure and input features are both generated using the same  resources,~\emph{i.e.},~attributes, which is more suitable for the use of a hypergraph.

\mycolor{As described above, the 20-newsgroup dataset adopted in our work does not have a pre-defined graph structure. To enable a direct comparison between GCN and hypergraph convolution feasible, we first need to construct the graph. For the graph construction of GCN, we connect two vertices if there is at least one shared word. In this case, we do not necessarily distinguish how many words they share. However, it is quite likely that more than two articles share the same word, thus naturally forming a hypergraph structure. Hence, we use words directly as hyperedges to connect articles for evaluating hypergraph convolution.}


\mycolor{As for the implementation, we use the same model architecture, the same training strategy, the same loss function and the same set of hyper-parameters as those in Sec.~\ref{sec:exp_citation}. That means, except for the way of graph construction, all the rest settings remain the same. Without bells and whistles, our hypergraph convolution reports an accuracy of $61.7$ while GCN reports an accuracy of $57.0$. The performance gain $4.7$ solely comes from a better way of modeling the occurrence of words, which suggests the capability of our method in handling non-pairwise structures.}


\section{Conclusion} \label{sec:con}
In this work, we have contributed two end-to-end trainable operators to the family of graph neural networks,~\emph{i.e.},~hypergraph convolution and hypergraph attention. While most variants of graph neural networks assume pairwise relationships of objects of interest, the proposed operators handle non-pairwise relationships modeled in a high-order hypergraph. We theoretically demonstrate that some recent representative works,~\emph{e.g.},~graph convolution network~\cite{GCN} and graph attention network~\cite{GAN}, are special cases of our methods. Hence, our proposed hypergraph convolution and hypergraph attention are more flexible in dealing with arbitrary orders of relationships and diverse applications, where both pairwise and non-pairwise formulations are likely to be involved. Thorough experimental results with semi-supervised node classification demonstrate the efficacy of the proposed methods.

\mycolor{In real applications, the key to applying our method is to abstract the non-pairwise relationships (for defining hyperedges) from the data. For example, i) in retrieval systems, a keyword, being as a hyperedge, could connect more than two webpages; ii) in collaborative filtering, a product tag, being as a hyperedge, could connect more than two users; iii) in social networks, two individuals could be connected by their direct friendship, which is a pairwise relationship in a vanilla graph. However, if a post is commented on by multiple individuals, we could use the post as a hyperedge to connect those individuals. Besides, our method does not distinguish undirected hypergraphs from directed hypergraphs. That means it is capable of handling both cases since the incident structure can be determined similarly. The last remark is about the scalability of our method. We have presented how to efficiently implement our method in Sec.~\ref{sec:summary} to accommodate popular deep learning platforms. When dealing with billion or trillion scale data, one may resort to distributed systems like MapReduce for the sake of commercial use.}

There are still some challenging directions that can be further investigated. Some of them are inherited from the limitation of graph neural networks, such as training substantially deeper models with more than a hundred of layers~\cite{resnet}, handling dynamic structures~\cite{survey1,survey2}, batch-wise model training,~\emph{etc}. Meanwhile, some issues are directly related to the proposed methods in high-order learning. For example, although hyperedges are equally weighted in our experiments, it is promising to exploit a proper weight mechanism when extra knowledge of data distributions is accessible, and even, adopt a learnable module in a neural network then optimize the weight with gradient descent. The current implementation of hypergraph attention cannot be executed when the vertex set and the hyperedge set are from two heterogeneous domains. One possible solution is to learn a joint embedding to project the vertex features and edge features~\cite{simonovsky2017dynamic,schlichtkrull2018modeling} in a shared latent space, which requires further exploration.

Moreover, it is also interesting to plug hypergraph convolution and hypergraph attention into other variants of graph neural network,~\emph{e.g.},~MoNet~\cite{monti2017geometric}, GraphSAGE~\cite{hamilton2017inductive} and GCPN~\cite{you2018graph}, and apply them to other domain-specific applications,~\emph{e.g.},~3D shape analysis~\cite{boscaini2016learning,xiao1,xiao2}, visual search~\cite{xiao3}, visual question answering~\cite{narasimhan2018out}, chemistry~\cite{gilmer2017neural,NIPS2018_8005}, knowledge graphs~\cite{Knowledge_Hypergraphs}, matrix factorization~\cite{jin2015low} and NP-hard problems~\cite{li2018combinatorial}.

\section*{Acknowledgment}
This work was supported by EPSRC grant Seebibyte EP/M013774/1 and EPSRC/MURI grant EP/N019474/1.

\bibliography{mybibfile}

\end{document}